\documentclass[11pt,a4paper]{article}
\usepackage[hyperref]{emnlp2020}

\usepackage{tabu} 
\usepackage{bbm}
\usepackage{transparent}
\usepackage{CJKutf8}
\usepackage{balance}

\usepackage{rotating}

\usepackage{url}
\usepackage{times}
\usepackage{latexsym}

\usepackage{url}
\usepackage{proof}
\usepackage{xspace}
\usepackage{multirow}
\usepackage{forest}
\usepackage{caption}
\usepackage{rotating}
\usepackage{dsfont}
\usepackage{amsmath}
\usepackage{amssymb}
\usepackage{amsthm}
\usepackage{algorithm}
\usepackage{float}
\usepackage{tikz}
\usepackage{fontawesome}
\usepackage{subfigure}
\usetikzlibrary{plotmarks}

\newcounter{ddaggerfootnote}

\newcommand{\namecite}[1]{\newcite{#1}}
\usepackage{bm}
\definecolor{ao_english}{rgb}{0.0, 0.5, 0.0}

\newcommand{\notes}[1]{}

\theoremstyle{definition}

\theoremstyle{plain}

\newcommand{\ith}[1]{\ensuremath{i^{{th}}}}

\newcount\permx
\newcount\permy
\def\permdot#1#2{
\permx=#1 \advance\permx by-1
\permy=#2 \advance\permy by-1
\psframe[fillcolor=black, fillstyle=solid]
(\permx,\permy)(#1, #2)
}

\newcommand{\boxnum}[1]{{\setlength{\fboxsep}{1pt}\raisebox{1pt}{\hspace{1pt}\fbox{\tiny #1}\hspace{1pt}}}}
\newcommand{\ind}[1]{\ensuremath{_{\kern-0.5pt\boxnum{#1}}}}

\newcommand{\vecx}{\ensuremath{\bm{x}}\xspace}
\newcommand{\vecy}{\ensuremath{\bm{y}}\xspace}

\def\namecite{\newcite}

\newcommand{\smallnt}[1]{\ensuremath{_{\mbox{\tiny PP}}}\xspace}

\newcommand{\pseudocode}{Algorithm}
\floatname{algorithm}{\pseudocode}

\iffalse

\else

\fi

\newcommand{\floor}[1]{\lfloor #1 \rfloor}
\newcommand{\gwaitk}{\ensuremath{{g_\text{wait-$k$}}}\xspace}
\newcommand{\gcatchup}{\ensuremath{{g_\text{wait-$k$, $c$}}}\xspace}

\definecolor{chocolate}{rgb}{0.28, 0.02, 0.03}
\definecolor{PaleGreen}{rgb}{0.33, 0.545,0.33}
\definecolor{colorC0}{RGB}{51,113, 169}
\definecolor{colorC1}{RGB}{243,130,37}
\usepackage{array}
\usepackage{diagbox}

\definecolor{dollarbill}{rgb}{0.52, 0.73, 0.4}
\definecolor{deepmagenta}{rgb}{0.8, 0.0, 0.8}
\definecolor{coralred}{rgb}{1.0, 0.25, 0.25}

\aclfinalcopy 

\title{Fluent and Low-latency Simultaneous Speech-to-Speech Translation \\
with Self-adaptive Training \thanks{\quad See our speech-to-speech simultaneous translation demos (including comparison with human interpreters) at  \scriptsize\url{https://sat-demo.github.io}.} }

\author{Renjie Zheng $^{1}$ \thanks{\quad Equal contribution} \quad
  Mingbo Ma $^{1\;\dagger}$ \quad
  Baigong Zheng $^{1}$ \thanks{\quad Work done at Baidu Research. Current address: Kwai Inc., Seattle, WA, USA. } \quad
  Kaibo Liu $^{1}$ \quad \\
  {\bf
  Jiahong Yuan $^{1}$ \quad
  Kenneth Church $^{1}$ \quad
  Liang Huang $^{1,2}$
  }
\\
  $^{1}$Baidu Research, Sunnyvale, CA, USA \\
  $^{2}$Oregon State University, Corvallis, OR, USA \\
  \texttt{\{renjiezheng, mingboma\}@baidu.com} \\
}

\date{}

\begin{document}

\begin{CJK}{UTF8}{gbsn}
\maketitle
\begin{abstract}

Simultaneous speech-to-speech translation is widely useful but extremely challenging,
since it needs to generate target-language speech {\em concurrently} with 
the source-language speech, with only a few seconds delay.
In addition, it needs to continuously translate a stream of sentences, but all recent solutions 
merely focus on the single-sentence scenario.
As a result, current approaches  accumulate 
 latencies progressively
when the speaker talks faster, 
and introduce unnatural pauses 
when the speaker talks slower.
To overcome these issues, we propose Self-Adaptive Translation (SAT)
which flexibly adjusts the length of translations 
to accommodate different source speech rates.
At similar levels of translation quality (as measured by BLEU), 
our method generates more fluent target speech 
(as measured by the naturalness metric MOS)
 with substantially lower
latency than the baseline, in both Zh$\leftrightarrow$En directions. 
\end{abstract}

\section{Introduction}

Simultaneous speech-to-speech translation, which mimics the human interpreter's practice to 
translate the source speech into a different language
with 3 to 5 seconds delay,
has wide usage scenarios such as international conference meetings, traveling and
negotiations
as it
provides more natural communication process than 
simultaneous speech-to-text translation.
This task has been widely considered as one of the most challenging tasks in NLP 
with (but not limited to) following reasons:
on one hand, the simultaneous translation is a hard task due to the
word order difference between source and 
target languages, e.g., SOV languages (German, Japanese, etc.) 
and SVO languages (English, Chinese, etc.);
on the other hand,  simultaneous speech-to-speech translation escalates
the challenge by considering 
the smooth cooperation between the modules of speech recognition, 
translation and speech synthesis.


\begin{figure}[!t]
\centering
\includegraphics[width=7.cm]{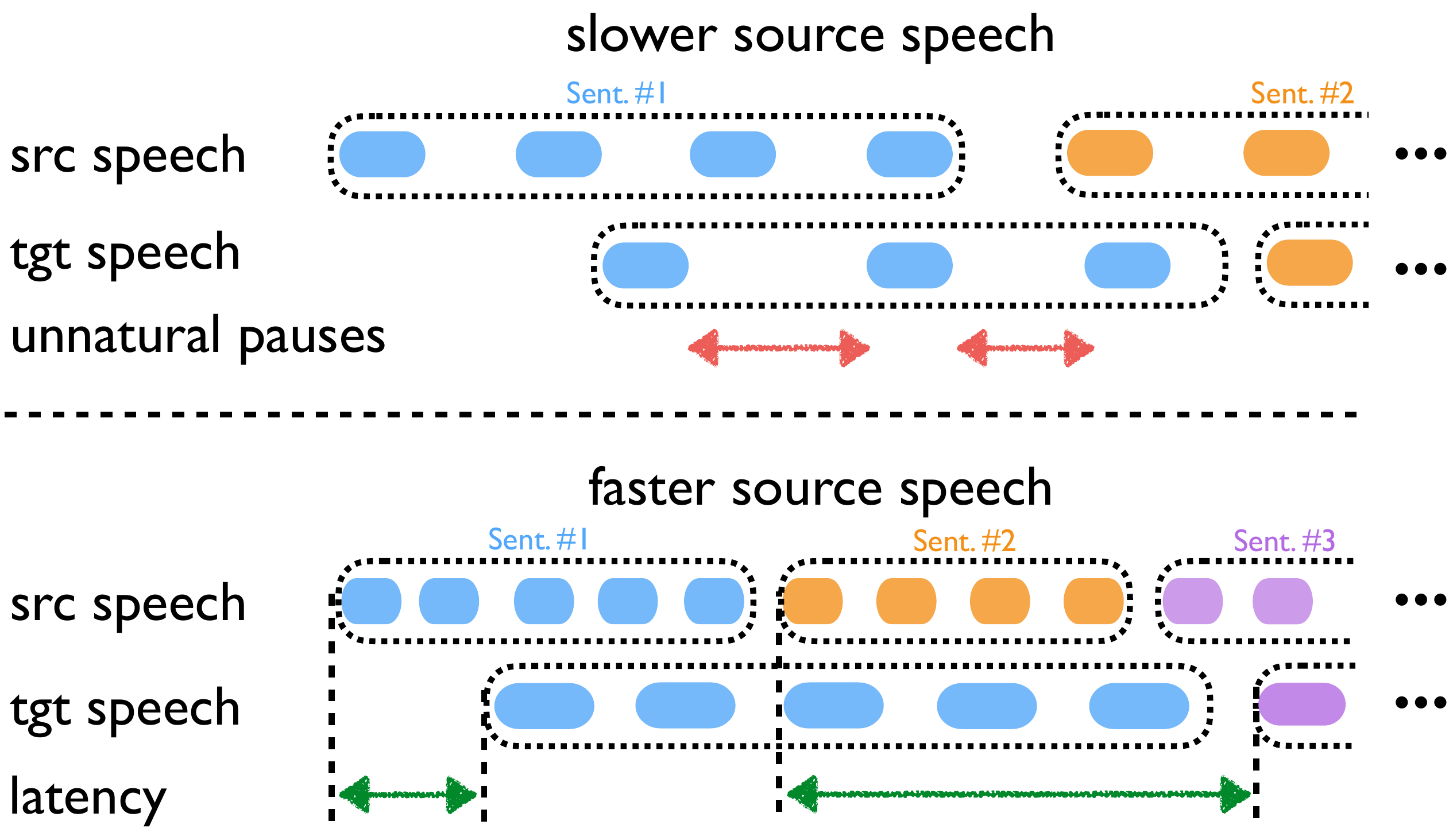}
\captionof{figure}{
Slower source speech causes unnatural pauses 
({\color{red}$\leftrightarrow$}) between words.
Faster source speech propagates extra latencies
({\color{ao_english}$\leftrightarrow$})
to the following sentences.
}
\label{fig:gaps_latency}
\vspace{-0.6cm}
\end{figure}

In order to achieve simultaneous speech-to-speech translation (SSST),
to the best of our knowledge,
most recent approaches \cite{oda+:2014,xiong+:2019} dismantle 
the entire system into a three-step pipelines, streaming Automatic Speech Recognition (ASR)
\cite{sainath+:2020,hirofumi+:2020,li+:2020},
simultaneous Text-to-Text translation (sT2T) \cite{Gu+:2017,ma+:2019,Arivazhagan+:2019,ma2019monotonic},
and Text-to-Speech (TTS) synthesis \cite{wang+:2017,ping+:2017,oord+:2018}.
Most recent efforts mainly focus on sT2T  which is considered  
the key component to further reduce the translation latency and improve the translation quality 
for the entire pipeline.
To achieve better translation quality and lower latency, 
there has been extensive research efforts which concentrate on the sT2T by introducing 
more robust models \cite{ma+:2019,Arivazhagan+:2019}, 
better policies \cite{Gu+:2017,zheng+:2020,zheng+:2019,zheng+:2019b},
new decoding algorithms \cite{zheng2019speculative,zheng+opportunistic:2020},
or multimodal information \cite{imankulova+:2019}. 
However, is it sufficient to only consider the effectiveness of sT2T and ignore
the interactions between other different components?

\begin{figure}[t]
\vspace{-0.4cm}
\centering
\includegraphics[width=6.5cm]{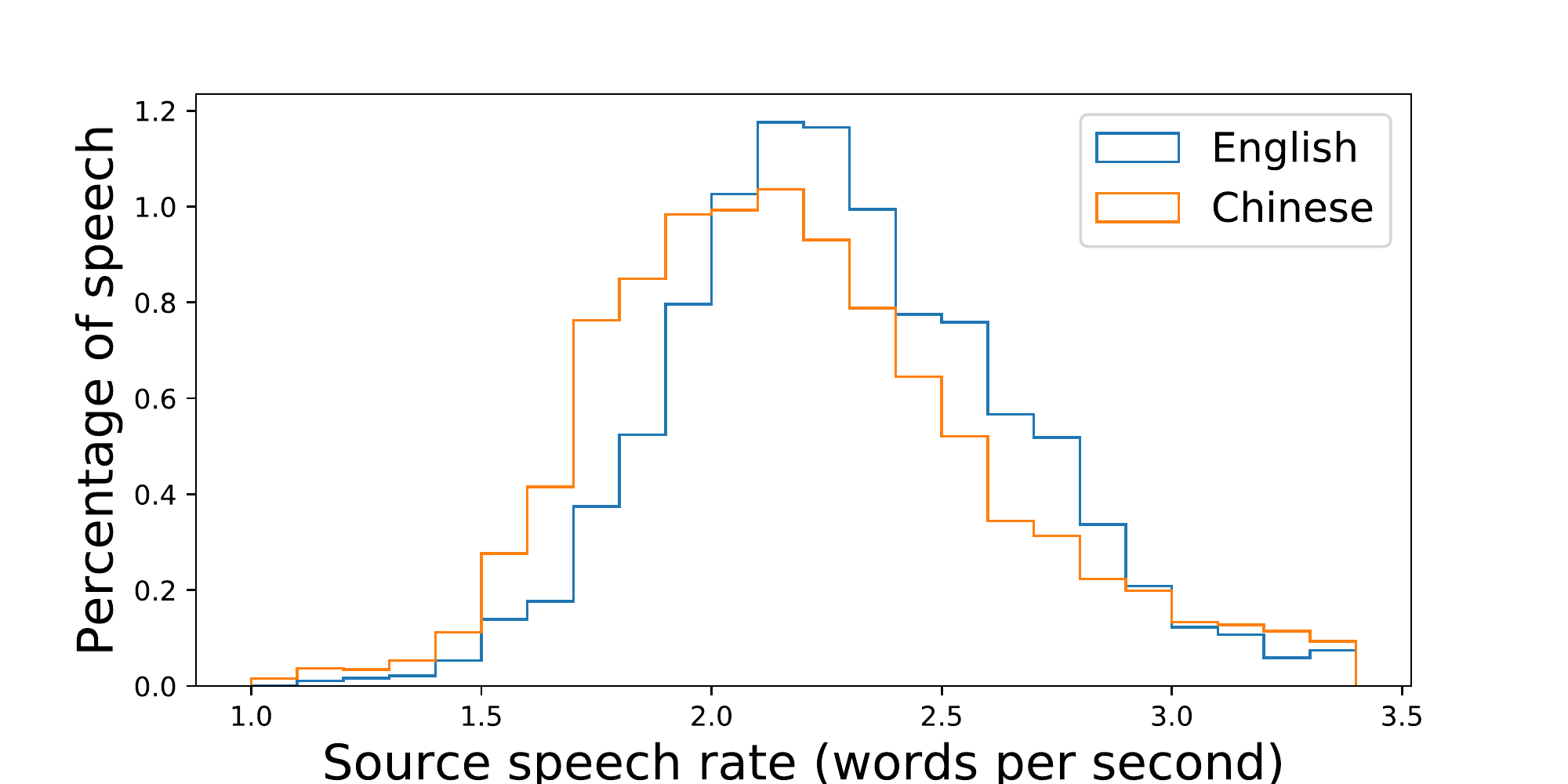}
\captionof{figure}{Average Chinese and English speed rate distribution for different speakers.}  
\label{fig:word_duration}
\vspace{-0.6cm}
\end{figure} 

Furthermore, in practice, when we need to translate multiple sentences continuously,
it is not only important to consider the cooperations between 
sT2T and other speech-related modules,
but also essential to take 
the effects between current and later sentence into consideration as 
shown in Fig.~\ref{fig:gaps_latency}.
Unfortunately, all the aforementioned techniques ignore other speech modules and 
merely establish their systems and analysis on
single-sentence scenario, which is not realistic.
To achieve fluent and constant, low-latency SSST, we also need to consider
speech speed difference between target and source speech.

\begin{figure}[t]
\centering
\vspace{-0.6cm}
\includegraphics[width=7.5cm]{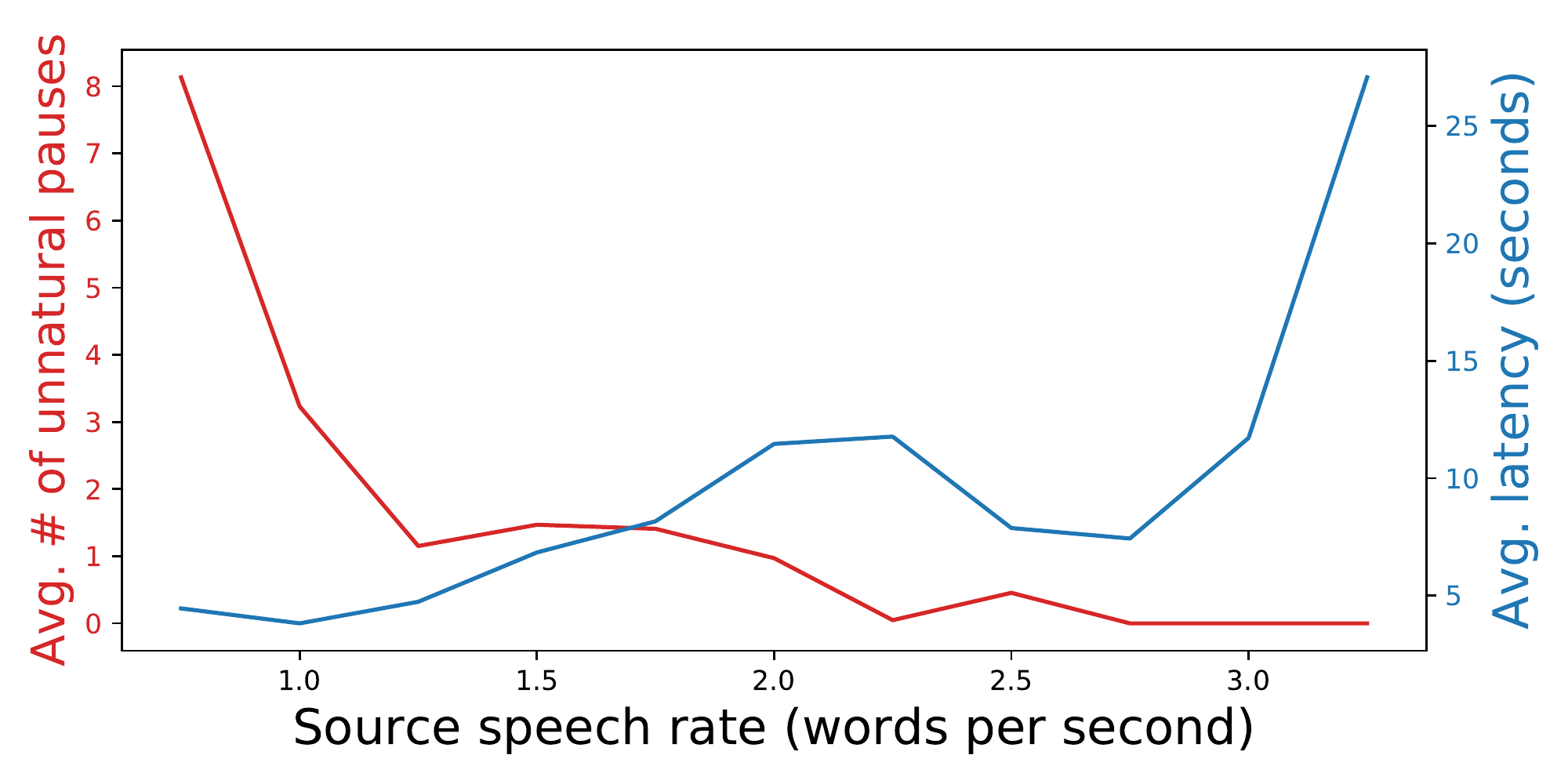}
\vspace{-0.2cm}
\captionof{figure}{Relationship between source speech rate with 
latency and number of unnatural pauses by naively using wait-$k$ model in
simultaneous Chinese-to-English speech-to-speech translation in our dev set.}
\label{fig:gaps_latency_exp}
\vspace{-0.6cm}
\end{figure}

Fig.~\ref{fig:word_duration} shows the speech rate distributions for 
both Chinese and English
in our speech corpus.
The speech rate varies especially for different speakers.
As shown in Fig.~\ref{fig:gaps_latency},
when we have various source speech speed,
the number of unnatural pauses and the latency vary 
dramatically. 
More specifically, when speaker talks slowly,
TTS often needs to make more pauses to wait for more tokens from sT2T which usually
does not output new translations with limited source information.
These unnatural pauses lead to semantic and syntactic 
confusion \cite{Lege2012TheEO,Bae}.
On the contrary, 
when speaker talks fast,
the target speech synthesized from previous sT2T models
(e.g. wait-$k$) always introduce large latency which
accumulates through the entire paragraph and 
causes significant delays.
Therefore, in realistic, the latency for the latter sentences are 
far more than the claimed latency in the original system. 
Fig.~\ref{fig:gaps_latency_exp} supports the above hypothesis
when source side speech rate varies
while using one wait-$k$ translation model and iTTS model.


\if
In practice, it is quit common to have inappropriate TSSR,
which could be cased by different TSTR, or different speech rate 
(number of words per second)
between source and target side.
As shown in Fig.~\ref{fig:gaps_latency},
when the TSSR is larger or smaller,
the number of unnatural pauses and the latency vary 
dramatically. 
More specifically, when TSSR is small,
TTS often need to make more pauses to wait for more tokens from sT2T which 
does not have new words to translate.
These unnatural pauses leads to semantic and syntactic 
confusion \cite{Lege2012TheEO,Bae}.
On the contrary, 
when TSSR is large,
the audio of previous translated sentence takes up the audio time of following sentence,
the latency from the earlier sentences accumulates through the entire paragraph and 
causes significant delays to the later sentences. 
Therefore, in realistic, the latency for the latter sentences are 
far more than the claimed latency in the original system. 
Fig.~\ref{fig:gaps_latency_exp} supports the above hypothesis
when we change the source side speech rate 
while keep fixing the 
number of words of both source and target side and target side speech rate.
\fi



To overcome the above problems, we propose Self-Adaptive Translation (SAT) for simultaneous 
speech-to-speech translation, which flexibly determines the length of translation
based on different source speech rate.
As it is shown in Figure~\ref{fig:comparison}, within this framework, 
when the speakers talk very fast, 
the model is encouraged to generate abbreviate but informative translation.
Hence, as a result of shorter translation, the previous
translation speech can
finish earlier and alleviate their effects to the latter ones.
Similarly, when the speakers have slower speech rate, 
the decoder will generate more meaningful tokens until a 
natural speech pause.
The speech pauses can be understood as a natural boundary between 
sentences or phrase which does not introduce ambiguity to the translation. 
In conclusion, 
we make the following contributions:

\begin{figure*}[t]
\centering
\includegraphics[width=15cm]{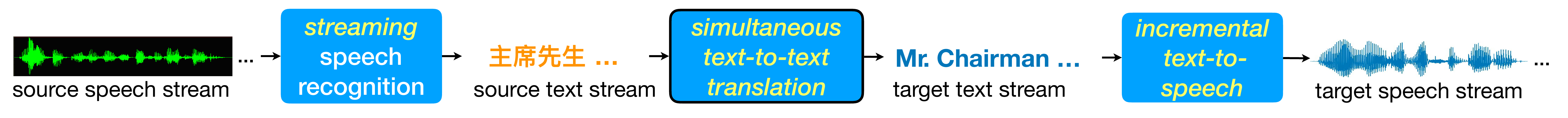}\vspace{-0.2cm}
\captionof{figure}{Pipeline of Speech-to-speech simultaneous translation.}
\label{fig:pipeline}
\vspace{-0.3cm}
\end{figure*}

\begin{itemize}
\setlength{\itemsep}{1pt}
	\item We propose SAT to flexibly adjust translation length  to generate fluent and low-latency target speeches for the entire speech (Sec.~\ref{sec:train}).
	\item We propose paragraph based Boundary Aware Delay as the first latency metric  suitable for simultaneous speech-to-speech translation 
	(Sec.~\ref{sec:metric}).
	\item We annotate a new simultaneous speech-to-speech translation dataset for Chinese-English translation, together with professional interpreters' interpretation (Sec~\ref{sec:exps}).
	\item Our system is the first simultaneous speech-to-speech translation system using iTTS (as opposed to full-sentence TTS) to further reduce the latency (Sec.~\ref{sec:prelim}). 
	\item Experiments show that our proposed system can achieve higher speech fluency and lower latency with similar or even higher translation quality compared with baselines and even human interpreters (Sec.~\ref{sec:exps}).
\end{itemize}




\vspace{-0.2cm}
\section{Preliminaries}
\vspace{-0.2cm}
\label{sec:prelim}

\begin{table}[t]\centering
\small
\begin{tabular}{|c|l|c|}\hline
\# & Incremental Transcription     &  Time (ms)    \\\hline
1 & thank you & 960 \\\hline
2 & thank you {\color{red}miss}  & 1120 \\\hline
3 & thank you {\color{blue}Mr} {\color{red}chair}  & 1600 \\\hline
4 & Thank you {\color{blue} ,} Mr {\color{blue}chairman .}  & 2040 \\\hline
\end{tabular}
\vspace{-0.1cm}
\caption{Example of English streaming ASR. Red words are
revised in latter steps. Punctuations only appear in the final step.}
\vspace{-0.4cm}
\label{tab:Zh-En}
\end{table}

In this section, we first introduce each component of three-step pipeline, which are streaming ASR, 
simultaneous translation models and incremental TTS techniques. 

\vspace{-0.2cm}
\subsection{Streaming Automatic Speech Recognition}
\label{sec:asr}

We use anonymous real-time speech recognizer as the speech recognition module.
As shown in Fig.~\ref{fig:pipeline}, streaming ASR is first step of the entire pipeline which 
converts the growing source acoustic signals from speaker into a sequence of 
tokens $\vecx=(x_1, x_2, ...)$ timely with about 1 second latency.
Table~\ref{tab:Zh-En} demonstrates one example of English streaming ASR 
which generates the English outputs incrementally.
Each row in the table represents the streaming ASR outputs at each step.
Note that streaming ASR sometimes revises some tail outputs from 
previous step (e.g. $3$th and $4$th steps in Table~\ref{tab:Zh-En}).
To get stabler outputs,
we exclude the last word in ASR outputs 
(except the final steps)
in our system.


\vspace{-0.2cm}
\subsection{Simultaneous Machine Translation}
\vspace{-0.2cm}

As an intermediate step between source speech recognition and target speech synthesis modules,
the goal of this step is to translation all the available source language tokens from streaming ASR
into another language.
There are many Text-to-Text simultaneous translation models \cite{Gu+:2017,ma+:2019,Arivazhagan+:2019,ma2019monotonic}
that have been proposed recently.

Different from conventional full-sentence translation model, 
which encodes the entire source sentence $\vecx=(x_1,...x_m)$ into a sequence of 
hidden states, and decodes sequentially conditioned on those hidden states and previous predictions
as  $p(\vecy \mid \vecx) = \textstyle\prod_{t=1}^{|\vecy|}  p(y_t \mid \vecx,\, \vecy_{<t})$
to form the final hypothesis $\vecy = (y_1,...,y_t)$, simultaneous translation 
makes predictions with partial, growing inputs before the source sentence finishes.

Without loss of generality, regardless the 
actual design of translation policy,
simultaneous translation can be represented
with prefix-to-prefix fashion as follows:
\vspace{-0.2cm}
\begin{equation}
p_g(\vecy \mid \vecx) = \textstyle\prod_{t=1}^{|\vecy|}  p(y_t \mid \vecx_{\leqslant g(t)},\, \vecy_{<t})
\label{eq:gensentscore2}
\vspace{-0.1cm}
\end{equation}
where $g(t)$ can be used to represent any arbitrary fixed or 
adaptive policy, denoting the number of processed source tokens at time step $t$.
We choose the wait-$k$ policy \cite{ma+:2019} 
as our baseline
for its simplicity and great performance.

More specifically, in this paper, our wait-$k$ policy is defined as follows:

\vspace{-0.2cm}
\begin{equation}
  \gwaitk(t) =\min\{k+t-1, \, |\vecx|\}
\label{eq:policy}
\vspace{-0.1cm}
\end{equation}
This policy starts to decode after the first $k$ source words, and 
then translates one token every time when one more source token
is received. 

\vspace{-0.1cm}
\subsection{Incremental Text-to-Speech}
\vspace{-0.1cm}

As the last step of the entire pipeline
the goal of iTTS is to incrementally generate
the target speech audio and play it to the audience
instantly with available translated words. 
Different from conventional full-sentence TTS, which requires the availability of the entire 
sentence, iTTS usually has 1-2 words delay but with similar audio quality compared with the
full-sentence TTS.
Compared with previous source sentence segment-based SSST systems \cite{oda+:2014,xiong+:2019}, our system can achieve word-level latency.
We adapt the iTTS framework from \namecite{ma+:2019} to our pipeline to generate 
target speech audio with translated tokens $\vecy_t$ at $t$ time step.

\section{Self-Adaptive Translation}
\label{sec:train}

To overcome the above practical problems,
we propose Self-Adaptive Translation (SAT) technique to enable the ability of 
adjusting the length of the translation based on the demand of latency and fluency.
We first demonstrate the problem of one naive solution.
Then, we introduce our training framework and talk about how to apply this 
technique in practice during inference time.

\vspace{-0.2cm}
\subsection{Naive Solution is Problematic}
\label{sec:naive}

To alleviate the various speech rate problem,
one naive solution is to adjust the target side speech speed based
on the source speaker's speed.
However, as shown in Table~\ref{tab:speed}, this solution is problematic 
as it usually requires the audience to be more focus on 
the translated speech
when we speed up the speech rate on target side, 
and sometimes it will
disrupt the audiences' comprehension of the translation
\cite{gordon2014recognition}.
Similarly, slowing down the speech only creates overlong 
phoneme pronunciation which is unnatural and leads to confusion.

\begin{table}[!]\centering
\small
\begin{tabular}{|c|c|}\hline
Speech Rate   & MOS   \\\hline
$0.5 \times$   &  $2.00 \pm 0.08$ \\\hline
$0.6 \times$  &  $2.32 \pm 0.08$ \\\hline
$0.75 \times$   &  $2.95 \pm 0.07$ \\\hline
Original        &  $4.01 \pm 0.08$ \\\hline
$1.33 \times$   &  $3.34 \pm 0.08$ \\\hline
$1.66 \times$   &  $2.40 \pm 0.09$ \\\hline
$2.0  \times$   &  $2.06 \pm 0.04$ \\ \hline
\end{tabular}
\caption{Mean Opinion Score (MOS) evaluations
of naturalness for different speech speed changed by ffmpeg. 
Original English speeches are synthesized by
our incremental Text-to-speech system. }
\label{tab:speed}
\vspace{-0.3cm}
\end{table}

Inspired by human interpreters\cite{he+:2016,Raja+:2000} 
who often summarize the contexts in order to catch up
the speaker, or make wordy translation to wait the speaker,
the optimal translation model should be enable to adjust the length of translated
sentence to change the speech duration on target side 
to avoid further delays or unnatural pauses fundamentally.

\vspace{-0.2cm}
\subsection{Self-Adaptive Training}

\begin{figure}[bt!]
\centering
\includegraphics[width=7.5cm]{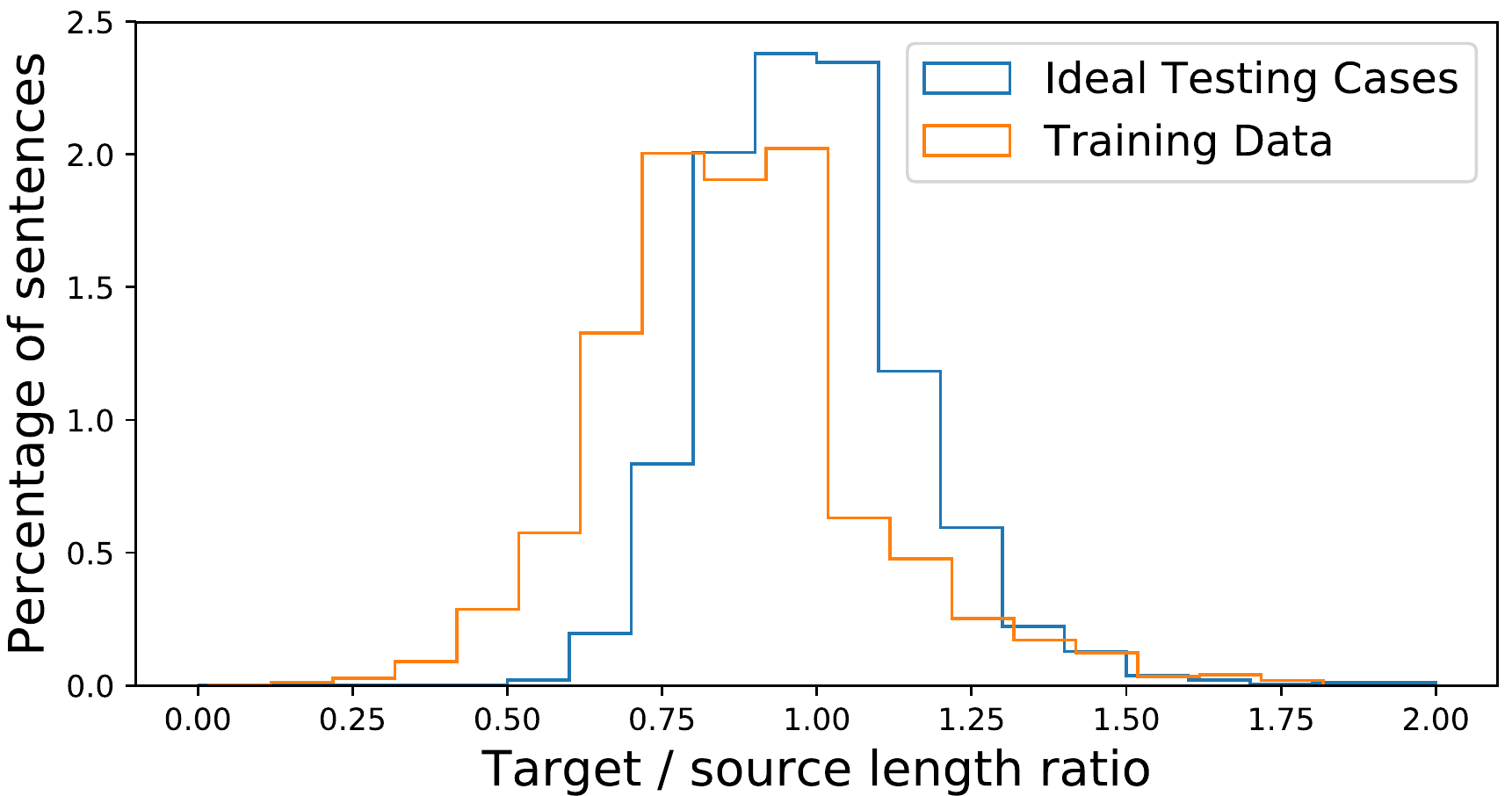}
\captionof{figure}{Tgt/src length ratio for English-to-Chinese task in training data (red) and ideal testing cases (blue).}  
\label{fig:lengthratio}
\vspace{-0.5cm}
\end{figure}

Translation between different language pairs
have various tgt/src length ratios,
e.g., English-to-Chinese translation ratio is roughly around 0.85
(small variations between different datasets).
However, this length ratio merely reflects the average statistics for the entire dataset,
and as it is shown with red line in Fig.~\ref{fig:lengthratio}, the ratio 
distribution for individual
sentence is quite wide around the average length ratio.

\begin{figure}[tb!]
\vspace{-0.5cm}
\centering
\includegraphics[width=7cm]{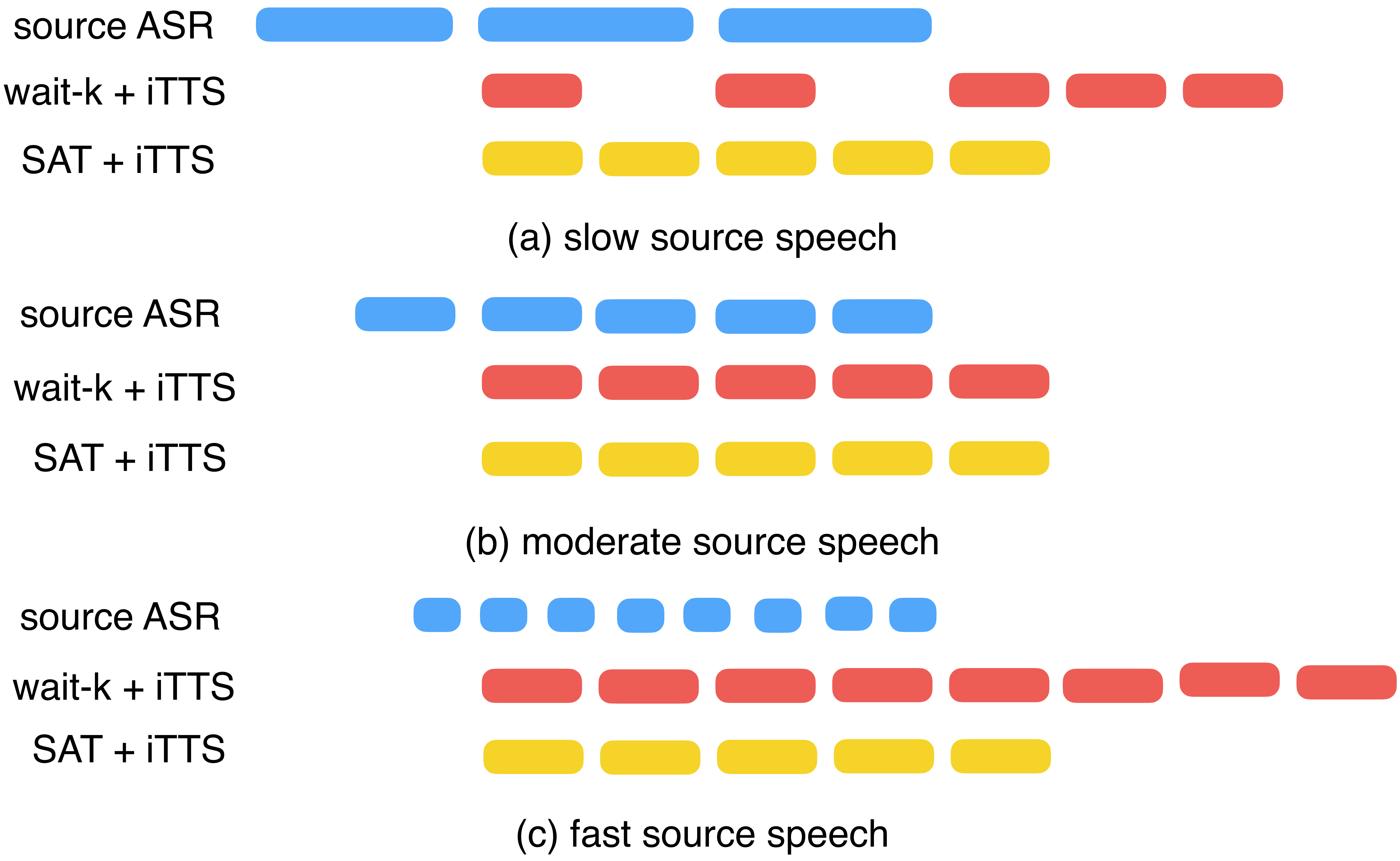}
\captionof{figure}{Illustration of conventional wait-$k$ (red) and SAT-$k$ (yellow) training policy. In SAT, we force the length of tail to be $k$ which equals the latency $k$. In the above example, we have $k=1$.}
\label{fig:comparison}
\vspace{-0.5cm}
\end{figure}

As shown in Fig.~\ref{fig:comparison} and discussed in earlier sections, 
over short and long translations are not preferred in simultaneous speech-to-speech
translation.
Ideally, we prefer the system 
to have a similar amount of 
initial wait 
with delay in the tail during translation of each sentence. 
Following this design, 
the translation tail of previous sentence
will fit perfectly into 
the beginning delay window for the following sentence,
and will 
not cause any  extra latency and intermittent speech.

Based on the above observation, we propose to use different training policies for different
sentences with different tgt/src ratios.
As shown in Fig.~\ref{fig:comparison},
We start from a fixed delay of $k$ tokens and 
then force the model to have the same
number of tokens in initial wait and final tail  
by amortizing the extra tokens into the middle steps.
More specifically, when we have longer tail than the fixed initial wait, 
we move extra words into former steps, and some steps before 
tail will decode more than one word at a time.
As a result, there will be some one-to-many policies
between source and target
and the model will learn to generate longer translations with shorter source text.
On the contrary, when we have shorter tail,
we perform extra reading on the source side and the
model will learn to generate 
shorter translation through this many-to-one policy.
Formally, we define our SAT training policy as follows:

\vspace{-0.6cm}
\begin{equation}
\gcatchup(t) = \min\{k+t-1-\floor{ct},\; |\vecx|\}
\label{eq:policyc}
\vspace{-0.2cm}
\end{equation}

where $c$ is the compensation rate which is decided by the tgt/src length ratio after deduction of $k$ tokens in
source initial and target tail.
For example, when tgt/src length ratio is 1.25, then 
$c={|tgt|-k\over|src|-k}-1 =1.25-1=0.25$, 
representing to decode 5 target
words for every 4 source words,
and model learn to generate wordy translation.
When target side is shorter than source side, $c$ becomes negative, and model learn to
decode less tokens than source side.

\begin{figure}[!h]
\centering
\includegraphics[width=5.8cm]{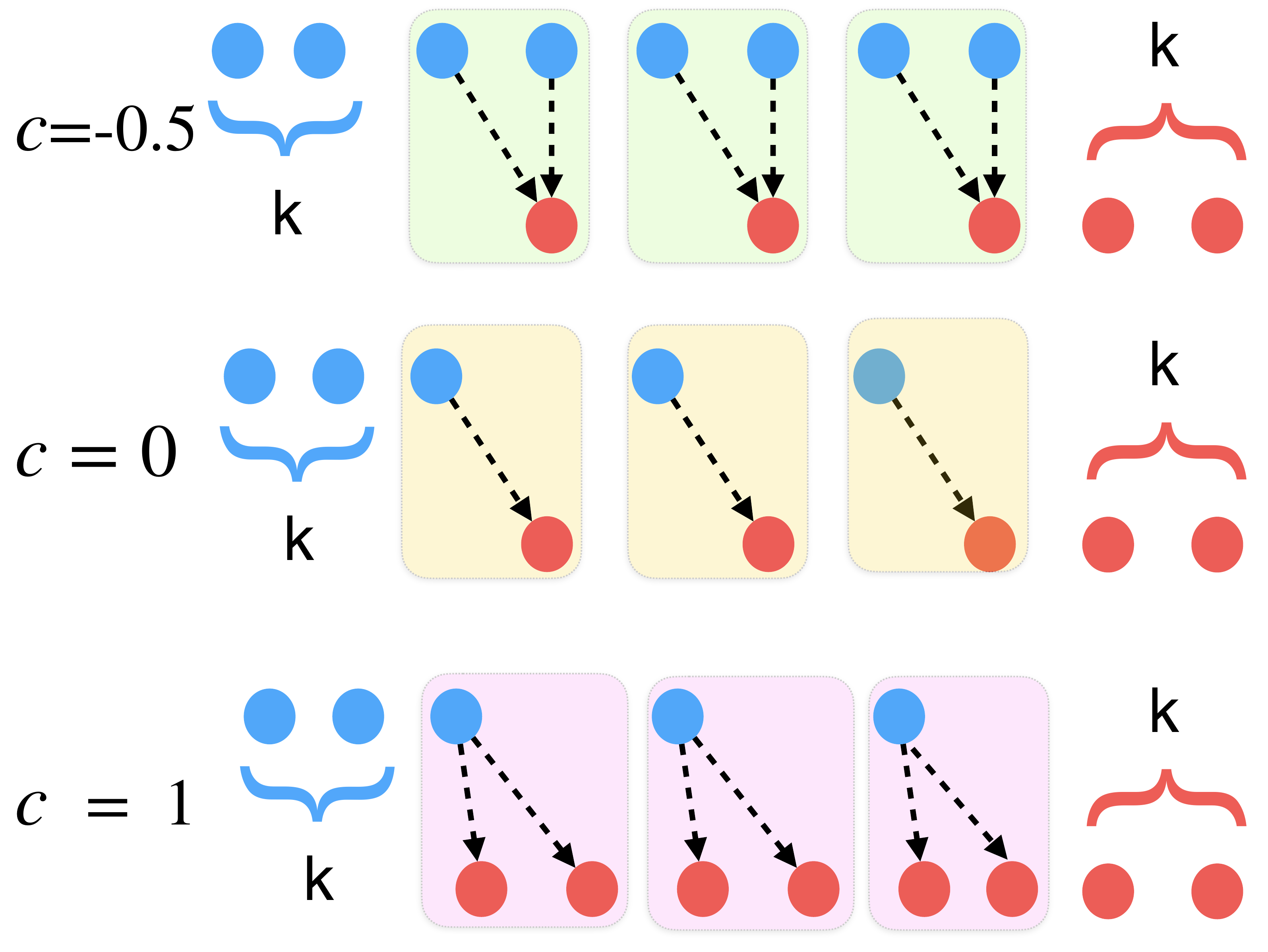}
\captionof{figure}{Different translation policy with different choice of $c$. Green boxes represent
many-to-1 policy; yellow boxes denote 1-to-1 policy; purple boxes show 1-to-many translation policy.}
\label{fig:mapping}
\vspace{-0.4cm}
\end{figure}

Note that the tgt/src length ratio in our case is determined by 
the corresponding sentence itself instead of the corpus level
tgt/src length ratio, which is a crucial different from 
catchup algorithm from \cite{ma+:2019}
where some short translations is trained with inappropriate positive $c$.
It seems to be a minor difference, but it actually enables the model to learn
totally different thing other than catchup.

The blue line in Fig.~\ref{fig:lengthratio} represents the tgt/src length ratio
for the ideal simultaneous speech-to-speech translation examples in our 
training set which have the same speech time between source and target side.
When we have the same speech time between source and target side,
there will be no accumulated latency from previous sentences to the following 
sentences.
As we notice, our training data covers all the tgt/src length ratio distribution
for the ideal cases, indicating that by adjusting the compensation rate $c$
from our training corpus, 
our model learns to generate appropriate length of translation on the target
side to avoid accumulated latency.

As shown in Fig.~\ref{fig:lengthratio}, there are many different choices of $c$ 
for different sentences, and each sentence is trained with their own
corresponding compensation rate which makes the training policy different from others with different $c$.
Hence, As shown in Fig.~\ref{fig:mapping},
our trained model is 
implicitly learned
many different policies, and when you choose 
a compensation rate $c$ during inference,
the model will generate certain length of translation corresponding to that 
compensation rate $c$ in training. 
More specifically, assume we have a source sentence, for example in Chinese, 
with length of $m$, and the 
conventional full-sentence or wait-$k$ model normally would translate this into
a English sentence with length of $1.25 \times m$.
However, the output length from SAT can be changed by $c$ 
following 
the policy in Eq.~\ref{eq:policyc} during decoding.
When $c$ is negative, SAT generates shorter translation than $1.25 \times m$.
On the contrary, if we choose $c$ that is positive, 
SAT generates longer translation than $1.25 \times m$.
The compensation rate $c$ functions as the key of model selection
to generate outputs of different lengths.

\begin{figure}[t]
\centering
\begin{tabular}{c}
\begin{minipage}[t]{1.0 \linewidth}
\vspace{-0.6cm}
\begin{center}
\subfigure[Tail length vs.~test-time compensation rate]{
\includegraphics[width=6.5cm]{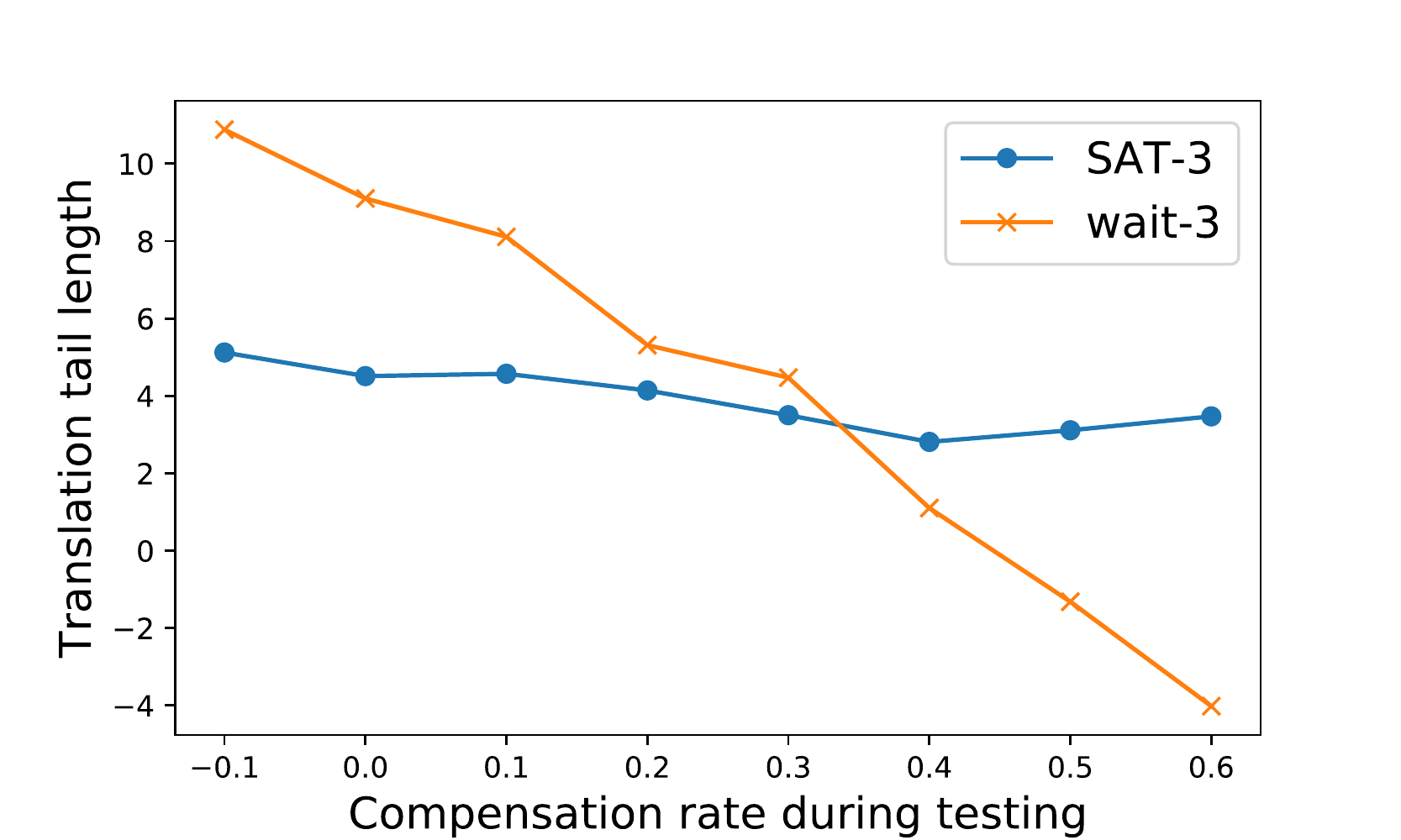}
\label{fig:tail_len}
\vspace{-0.6cm}
}
\end{center}
\end{minipage}
\vspace{-0.5cm}
\\
\vspace{-0.4cm}
\begin{minipage}[t]{1.0 \linewidth}
\begin{center}
\subfigure[Translation  length $|\vecy|$ vs.~test-time compensation rate]{
\includegraphics[width=6.5cm]{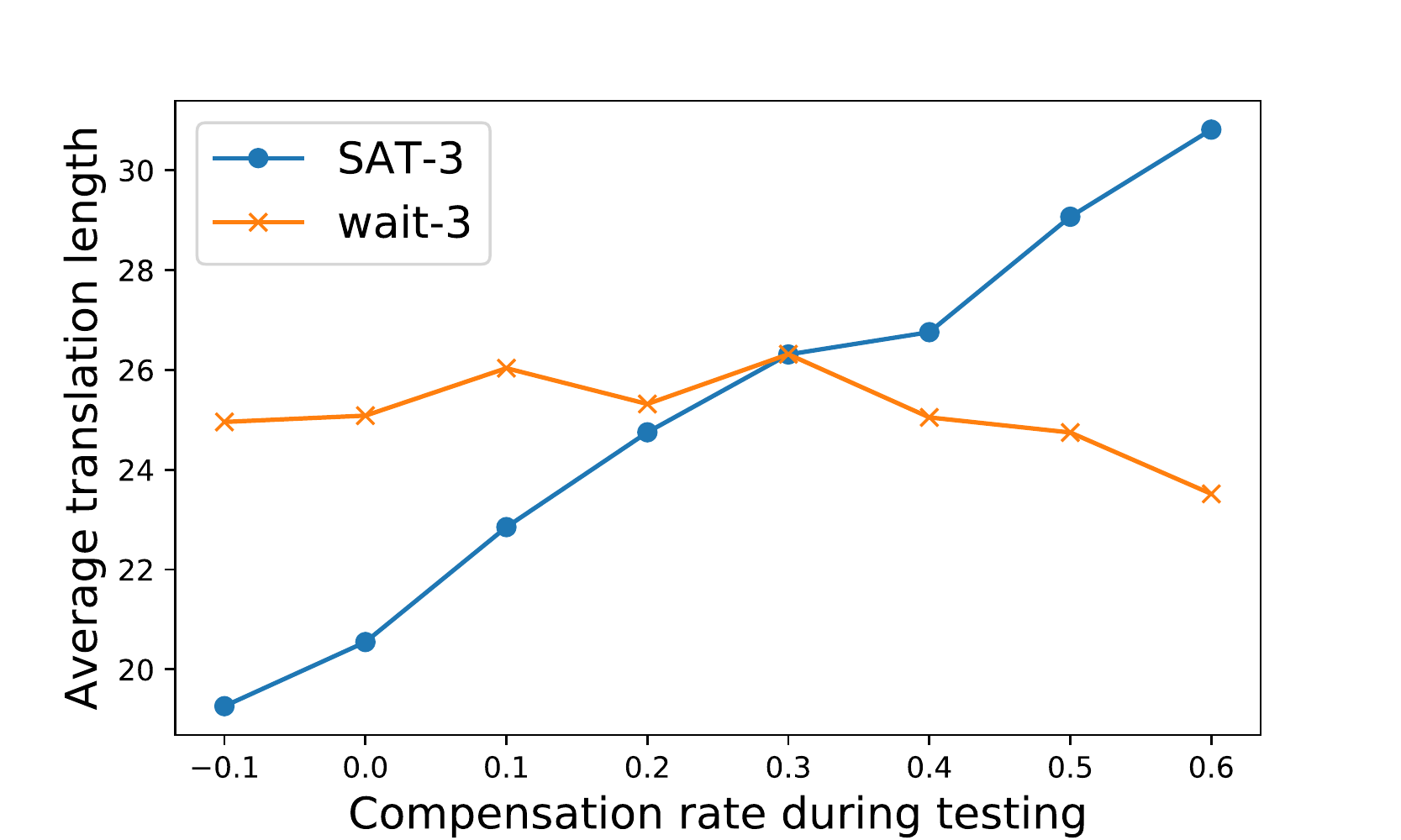}
\label{fig:tgt_len}
\vspace{-0.6cm}
}
\end{center}
\end{minipage}
\end{tabular}
\caption{Translation length analysis on Chinese-to-English
task using one SAT-$3$ and wait-$k$ model.}
\vspace{-0.4cm}
\end{figure}

Fig.~\ref{fig:tail_len}--\ref{fig:tgt_len} show the effectiveness of our proposed
model which has the ability to adjust the tail length of the entire translation
with different $c$'s.

\subsection{Self-Adaptive Inference}
\label{sec:SAI}

The above section discusses the importance of $c$, which is easily to obtain during training time,
but at inference time, we do not know the optimal choice of $c$ in advance
since the fluency and latency criteria also rely on the finish time for each word on both sides.
Therefore, the streaming ASR and iTTS plays important roles 
here to determine the decoding policy to form fluent and low-latency translation speech
and we use the knowledge of streaming ASR and iTTS for selecting the appropriate 
policy on the fly.

When we have a faster speech, 
streaming ASR will send multiple tokens to SAT at some steps.
But SAT only
generates one token at a time on the target side and pass it to iTTS instantly.
This decoding policy has a similar function to a negative  $c$, which
has many-to-one translation policy. 
In this case, SAT will generate succinct
translation and iTTS therefor can finish the translation speech with shorter time since
there are less tokens.

Contrarily, when the speaker talks with slower pace, there is only one token that is
feed into SAT. 
SAT translates it into a different token and delivers it to iTTS.
This is one-to-one translation policy.
When iTTS is about to finish playing the newly generated speech, and there is no new incoming
token from steaming ASR, 
SAT will force the decoder to generate one extra token, 
which becomes one-to-many translation policy 
(including the decoded token in previous step), 
and feeds it to iTTS.
When the speaker makes a long pause, and there is still no new tokens from streaming ASR,
the decoder of SAT continues to translate until a pause token (e.g. any punctuations) generated.
This pause token forms a necessary, natural pause on speech side and will not change 
the understanding for the translated speech.

\vspace{-0.2cm}
\section{Paragraph-Based Boundary Aware Latency}
\vspace{-0.2cm}
\label{sec:metric}

As mentioned frequently above, latency is another essential
dimension for simultaneous speech-to-speech translation performance.
However, the measurement of speech delay from source speech to each synthesized word in target speech is challenging and 
there is no direct metric that is suitable for simultaneous speech-to-speech translation.

Human interpreters use Ear-Voice-Span (EVS) \cite{gumul2006conjunctive,lee2002ear} to calculate translation delay 
for some landmark words from source speech to target source.
However, this requires the target-to-source word correspondence.
In practice, the translation model sometimes makes errors during translation which
includes miss translation of some words or over translated some words that source does not
include.
Thus, an automatic fully word-to-word alignment 
between target and source is hard to be accurate.
Human annotation is accurate but expensive and not practical.

Inspired by \namecite{ari+:2020} who proposed Translation Lag (TL) 
to ignore the semantic correspondence between 
words from target to source side and only calculate
each target delay proportionally 
to each source words regardless the actually meaning of word in the task of 
simultaneous ``speech-to-text'' translation,
we use a similar method to calculate the latency for each sentence.

Nevertheless, TL is only designed for single-sentence latency ,
while we need to
measure the latency of a paragraph of speech.
Thus, we propose paragraph based Boundary Aware Latency
(pBAL) to compute the latency 
of long speech simultaneous translation.
In pBAL, we first align the each sentence, make each word's correspondence within the
sentence boundary.
Then we compute the time differences of the finished time between each target word's audio 
and its proportion corresponding source word's finish time
in source side.
In experiments, we determine the finish time of
each source and target words by forced aligner \cite{yuan2008speaker}
and align the translation and source speech
by using the corresponding streaming ASR as a bridge.

\vspace{-0.2cm}
\section{Experiments}
\label{sec:exps}
\vspace{-0.1cm}
\subsection{Datasets and Systems Settings}
\vspace{-0.1cm}

We evaluate on two simultaneous
speech-to-speech translation directions:
Chinese$\leftrightarrow$English.
For  training, we use the text-to-text
 parallel corpora available
from WMT18\footnote{\scriptsize\url{http://www.statmt.org/wmt18/translation-task.html}}
(24.7M sentence pairs).
We also annotate a portion of Chinese and English speeches
from LDC United Nations Proceedings Speech
\footnote{\scriptsize\url{https://catalog.ldc.upenn.edu/LDC2014S08}} (LDC-UN)
as a speech-to-text corpus.
This corpus includes speeches recorded in 2009-2012
from United Nations conferences in six official UN
languages.
We transcribe the speeches and then translate 
the transcriptions as references.
The speech recordings include not only source speech
but also corresponding professional  simultaneous
interpreters' interpretation in the conference.
Thus, we also transcribe those human simultaneous
interpretation of En$\rightarrow$Zh direction
which will not be used in our model but compared to in 
the following experiments.

\begin{table}[h!]\centering
\small
\begin{tabular}{|c|c|c|c|}\hline
\multicolumn{2}{|c|}{}  & En$\rightarrow$Zh & Zh$\rightarrow$En    \\\hline
\multirow{3}{*}{Train}  & \# of speeches & 58  & 119 \\
             & \# of words &   63650        & 61676 \\
          & Total time         &   6.81 h    & 9.68 h \\\hline
\multirow{3}{*}{Dev}& \# of speeches   & 3 &  6\\
          & \# of words    &   1153        & 2415 \\
          & Total time      &  0.27   h      & 0.35 h \\\hline
\multirow{3}{*}{Test}      & \# of speeches  & 3 & 6 \\
          & \# of words    &    3053           & 1870 \\
          & Total time      &   0.39 h       & 0.30 h \\\hline
\end{tabular}
\caption{Statistics of  LDC-UN dataset (source-side).}
\vspace{-0.4cm}
\label{tab:data}
\end{table}

Table \ref{tab:data} shows the statistics of our 
speech-to-text dataset.
We train our models using both the WMT18 training set and
the LDC UN speech-to-text training set.
We validate and test the models only in the LDC-UN dataset.
For Chinese side text, we use jieba 
\footnote{\scriptsize\url{https://github.com/fxsjy/jieba}} 
Chinese segmentation tool.
We apply BPE~\cite{sennrich+:2015} on all texts in order 
to reduce the vocabulary sizes. 
We set the vocabulary size to 16K for both Chinese and English text.
Our Transformer is essentially the same with base Transformer model 
 \cite{vaswani+:2017}.

As mentioned in Section \ref{sec:asr}, we use an
anonymous real-time speech recognizer from a 
well-known cloud platform
as the speech recognition module
for both English and Chinese.
During speech-to-speech simultaneous translation decoding,
after receiving an ASR input, we first normalize the
punctuations and tokenize (or do Chinese segmentation
for Zh$\to$En translation) the input.
The last tokens are always removed in the encoder
of translation model because they are very unstable.
In the latency measurement we use 
Penn Phonetics Lab Forced Aligner (P2FA) \cite{yuan2008speaker} as the forced aligner to automatically
annotate the time-stamp for both Chinese and English words
in source and target sides.

For the incremental Text-to-speech system, we follow
\namecite{ma2019incremental} 
and take the Tacotron 2 model~\cite{shen+:2018} as our phoneme-to-spectrogram model and train it with additional {\em guided attention loss}~\cite{tachibana+:2018} which speeds up convergence.
Our vocoder is the same as that in the Parallel WaveGAN paper~\cite{yamamoto+:19}, which consists of 30 layers of dilated residual convolution blocks with exponentially increasing three dilation cycles, 64 residual and skip channels and the convolution filter size 3. 
For English, we use a proprietary speech dataset containing 13,708 audio clips (i.e., sentences) from a female speaker and the corresponding transcripts.
For Chinese, we use a public speech dataset\footnote{\url{https://www.data-baker.com/open_source.html}} containing 10,000 audio clips from a female speaker and the transcripts.

\vspace{-0.2cm}
\subsection{Speech-to-Speech Simul.~Translation}
\vspace{-0.2cm}

\begin{figure}[t]
\centering
\vspace{-.9cm}
\begin{tabular}{c}
\begin{minipage}[t]{1.0 \linewidth}
\begin{center}
\subfigure[Chinese-to-English simultaneous translation]{
\includegraphics[width=5.9cm]{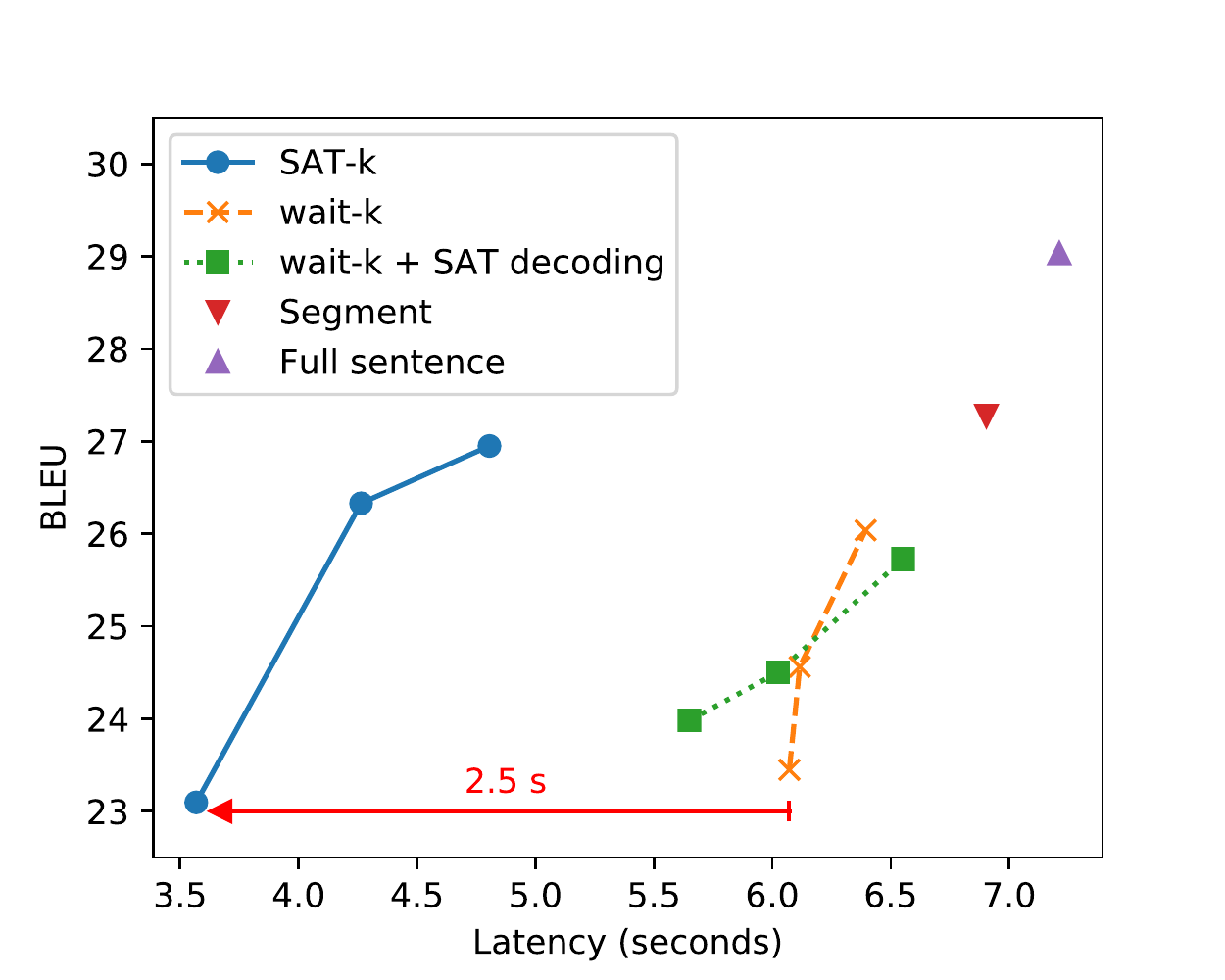}
\label{fig:zh-en}
}
\end{center}
\end{minipage}
\\[-0.3cm]
\begin{minipage}[t]{1.0 \linewidth}
\begin{center}
\subfigure[English-to-Chinese simultaneous translation]{
\includegraphics[width=6.5cm]{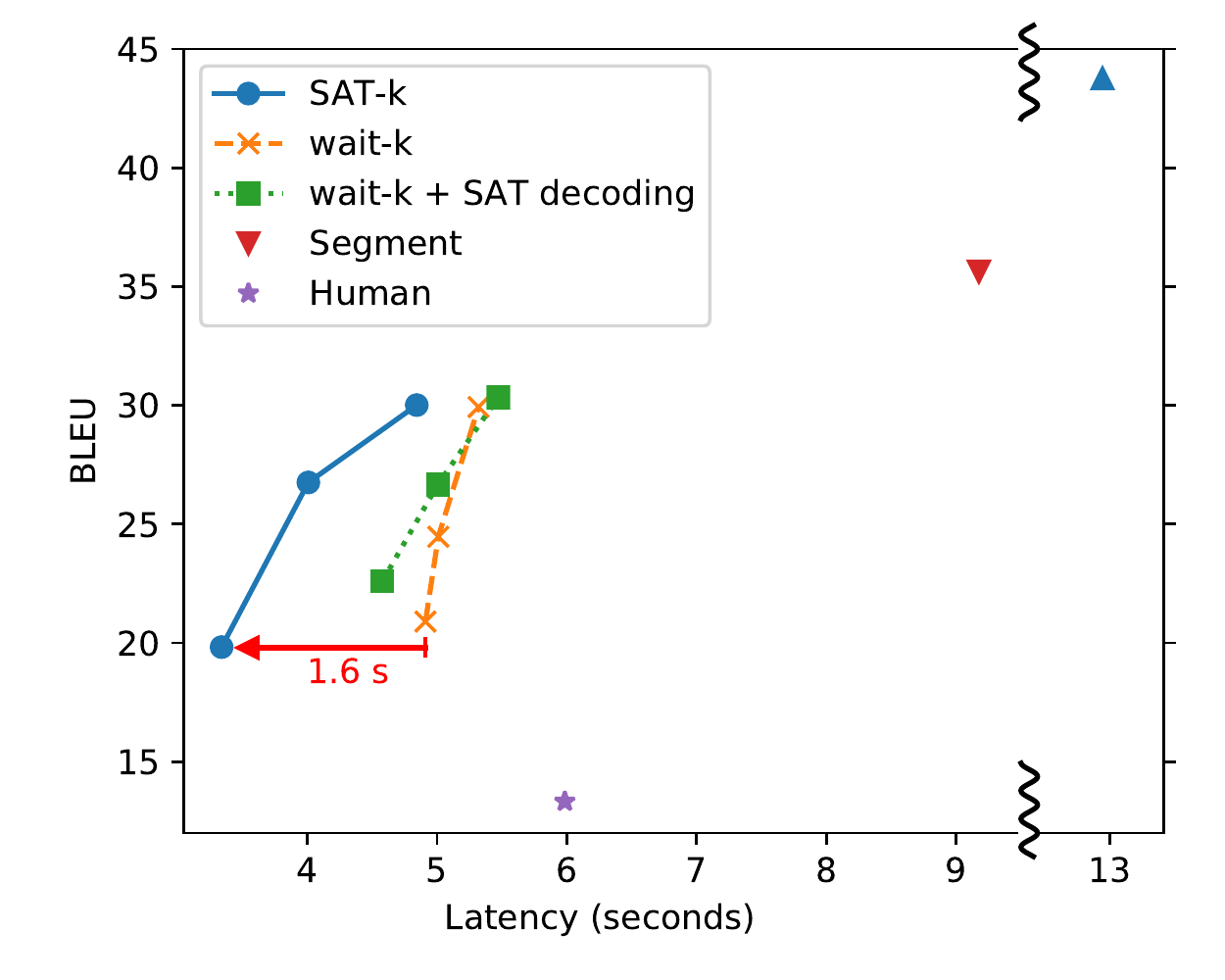}
\label{fig:en-zh}
}
\end{center}
\end{minipage}
\end{tabular}
\vspace{-0.3cm}
\caption{Translation quality and latency (pBAL)
of proposed simultaneous speech-to-speech translation
systems compared with baselines. For all those SAT-$k$ and wait-$k$ models, $k=\{3, 5, 7\}$ from bottom to top.}
\label{fig:result}
\vspace{-0.5cm}
\end{figure}

Fig.~\ref{fig:result} show the final results of our proposed 
models and baselines.
For translation quality measurement, we use the "multi-bleu.pl" 
\footnote{\scriptsize\url{https://github.com/moses-smt//mosesdecoder/blob/master/scripts/generic/multi-bleu.perl}}
script to calculate BLEU scores.
Since different punctuations are soundless,
we remove all of them before BLEU evaluation for both
hypotheses and references.
We follow \cite{xiong+:2019} to concatenate the translations
of each talk into one sentence to measure BLEU scores.

For Chinese-to-English simultaneous translation, 
we compare our models with naive wait-$k$,
wait-$k$ with SAT decoding (only use 
Self-adaptive inference in Sec.~\ref{sec:SAI}), 
segment based models \cite{oda+:2014,xiong+:2019} and full sentence translation model.
All these models share one iTTS system.
For segment based model,
since our streaming ASR API doesn't provide any punctuation
before the final step, 
we use the final punctuations to segment the partial 
streaming inputs and then use a full-sentence translation
model to translate the partial segment as a full sentence.
The results show that our proposed SAT-$k$ models can
achieve much lower latency without sacrificing  
quality compared with those baselines.

\begin{figure}[h!]
\centering
\vspace{-0.2cm}
\includegraphics[width=7.cm,height=4.cm]{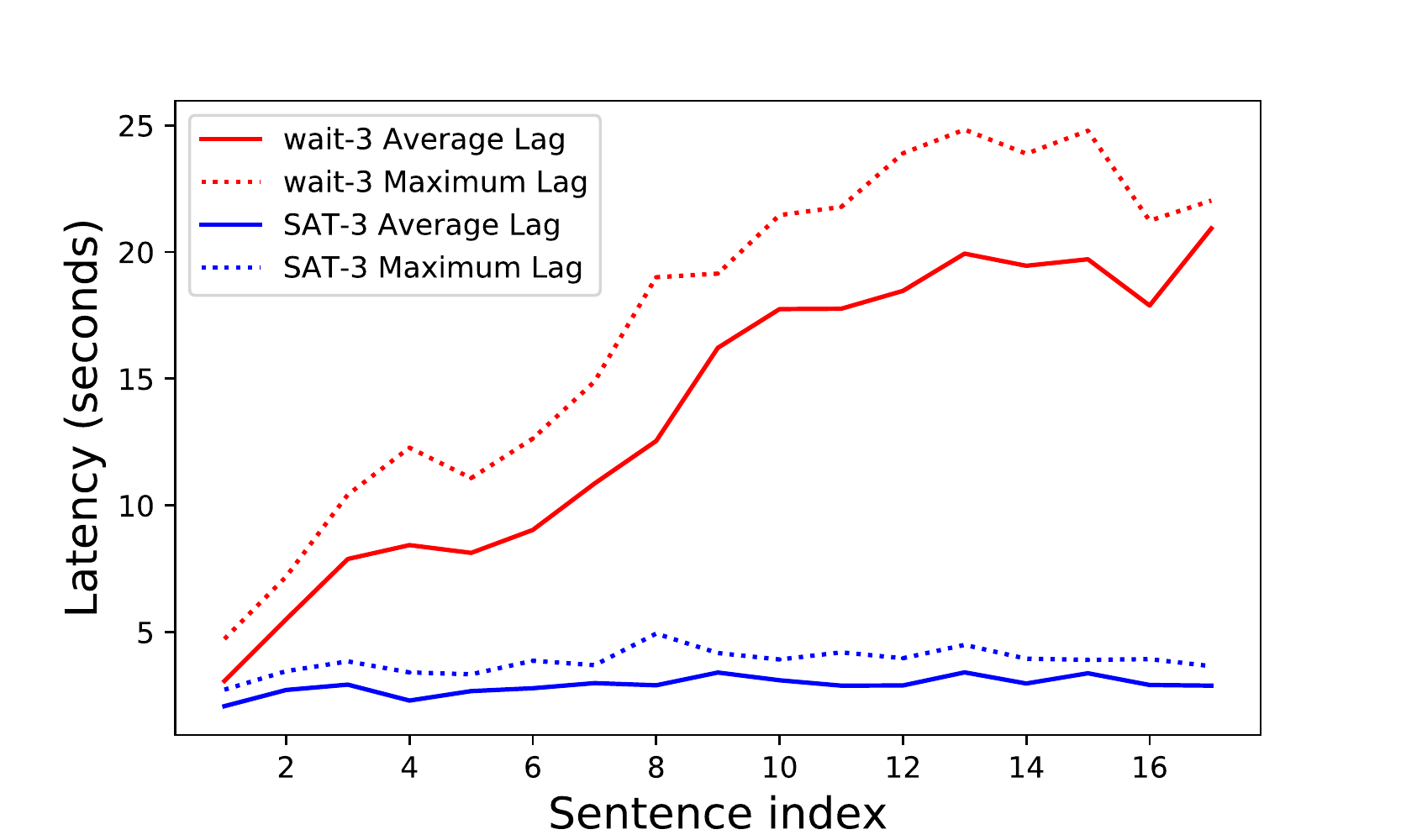}
\captionof{figure}{Latency for sentences at different 
indices in Chinese-to-English dev-set.}
\label{fig:acc_latency}
\vspace{-0.4cm}
\end{figure}

Fig.~\ref{fig:result}(b) shows the results of
En$\to$Zh simultaneous translation.
Besides the baselines used in Zh$\to$En
experiments, we also compare our system with
professional human interpreters' translation.
Our proposed models also outperform all the baselines
and human interpreters.
Our models reduce more latency in Zh$\to$En 
than En$\to$Zh compared
with wait-$k$ 
because
English sentences is always longer than Chinese
thus
it's more easily to accumulate
latency in Zh$\to$En 
(also shown in Fig.~\ref{fig:acc_latency}).

\vspace{-0.2cm}
\subsection{Human Evaluation on Speech Quality}
\vspace{-0.2cm}

\begin{table}[h!]\centering
\small
\begin{tabular}{|c|c|c|}\hline
Method                  &  En$\to$Zh & Zh$\to$En   \\\hline
wait-$3$                &  $3.56 \pm 0.09$ & $3.68 \pm 0.08$ \\\hline
wait-$3$ + SAT decoding  &  $3.81 \pm 0.08$ & $3.96 \pm 0.04$\\\hline
SAT-$3$             &  $3.83 \pm 0.07$ & $3.97 \pm 0.07$\\\hline
Segment-based           &  $3.79 \pm 0.15$ &$3.99 \pm 0.07$ \\\hline
Full sentence           &  $3.98 \pm 0.08$ &$4.03 \pm 0.03$ \\\hline
Human                   &  $3.85 \pm 0.05$ & - \\\hline
\end{tabular}
\caption{MOS evaluations
of fluency for different 
target speeches generated
by different methods.}
\label{tab:mos}
\vspace{-0.4cm}
\end{table}

In Table~\ref{tab:mos}, 
we evaluate our synthesized speeches by Mean Opinion Scores (MOS) 
with native speakers,
which is a standard metric in TTS.
Each speech received 10 human ratings scaled from 1 to 5,
with 5 being the best.
For both Zh$\leftrightarrow$En directions,
wait-$3$ models have the lowest MOS due to the many unnatural pauses 
(see Sec.~\ref{sec:naive}).
Our proposed model SAT-$3$ and wait-$3$
with SAT decoding achieve  similar
fluency to full sentence models and
even human interpreters.

\vspace{-0.2cm}
\subsection{Examples}

\begin{sidewaysfigure}
\centering
\begin{minipage}{1.0\textheight}
\includegraphics[width=25cm]{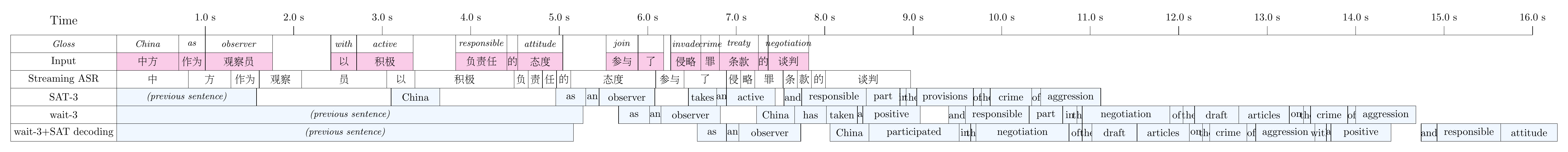}
\captionof{figure}{Decoding results of proposed simultaneous speech-to-speech Chinese-to-English translation system
and baselines.}
\label{fig:zhen_example}
\end{minipage}
\begin{minipage}{1.0\textheight}
\includegraphics[width=25cm]{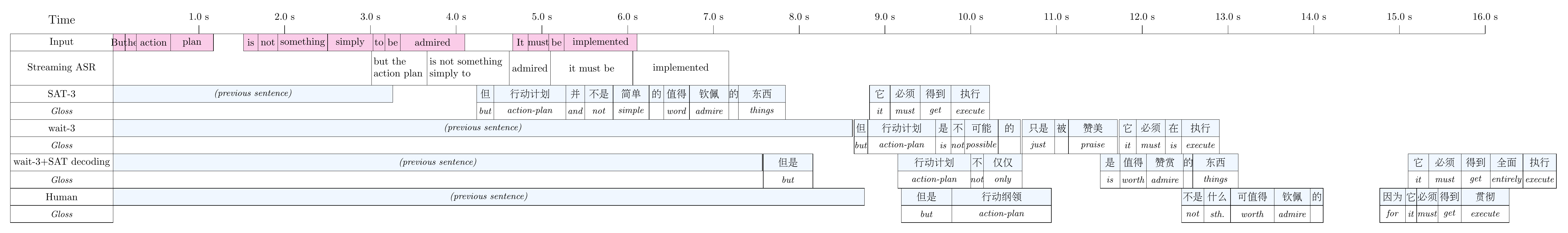}
\captionof{figure}{Decoding results of proposed simultaneous speech-to-speech English-to-Chinese translation system
and baselines.}
\label{fig:enzh_example}
\end{minipage}
\end{sidewaysfigure}

Fig.~\ref{fig:zhen_example} shows a Zh$\to$En 
decoding example. Here the wait-$3$ models'
outputs have a much longer latency compared with
SAT-$3$ because their beginnings are delayed by the translation of
previous sentence(s) and their tails are also very long.
The En$\to$Zh example in Fig.~\ref{fig:enzh_example} is similar.
While streaming ASR has a very long delay, 
SAT-$3$ model still controls the latency to roughly  4.5s;
all  pauses on the target side are natural ones from punctuations.
By contrast, the human interpreter's translation
has the longest latency.

\vspace{-0.2cm}
\section{Conclusions}
\vspace{-0.2cm}
We proposed Self-Adaptive Translation
for simultaneous 
speech-to-speech translation which flexibly adjusts translation length 
to avoid latency accumulation and unnatural pauses.
In both Zh$\leftrightarrow$En directions, our method generates fluent 
and low latency 
target speeches with high translation quality.

\balance

\bibliography{main}

\begin{thebibliography}{33}
\expandafter\ifx\csname natexlab\endcsname\relax\def\natexlab#1{#1}\fi

\bibitem[{Al-Khanji et~al.(2000)Al-Khanji, Shiyab, and Hussein}]{Raja+:2000}
Raja Al-Khanji, Said Shiyab, and Riyad Hussein. 2000.
\newblock \href {https://doi.org/10.7202/001873ar} {On the use of compensatory
  strategies in simultaneous interpretation}.
\newblock \emph{Meta: Journal des traducteurs}, 45:548.

\bibitem[{Ari et~al.(2020)Ari, Cherry, I, Macherey, Baljekar, and
  Foster}]{ari+:2020}
Naveen Ari, Colin~Andrew Cherry, Te~I, Wolfgang Macherey, Pallavi Baljekar, and
  George Foster. 2020.
\newblock \href {https://arxiv.org/abs/1912.03393} {Re-translation strategies
  for long form, simultaneous, spoken language translation}.
\newblock In \emph{ICASSP 2020}.

\bibitem[{Arivazhagan et~al.(2019)Arivazhagan, Cherry, Macherey, Chiu, Yavuz,
  Pang, Li, and Raffel}]{Arivazhagan+:2019}
Naveen Arivazhagan, Colin Cherry, Wolfgang Macherey, Chung-Cheng Chiu, Semih
  Yavuz, Ruoming Pang, Wei Li, and Colin Raffel. 2019.
\newblock Monotonic infinite lookback attention for simultaneous machine
  translation.
\newblock In \emph{Proceedings of the 57th Annual Meeting of the Association
  for Computational Linguistics}, Florence, Italy. Association for
  Computational Linguistics.

\bibitem[{Bae(2015)}]{Bae}
Rebecca~M. Bae. 2015.
\newblock \emph{The effects of pausing on comprehensibility}.
\newblock {PhD} dissertation, Iowa State University.

\bibitem[{Gordon-Salant et~al.(2014)Gordon-Salant, Zion, and
  Espy-Wilson}]{gordon2014recognition}
Sandra Gordon-Salant, Danielle~J Zion, and Carol Espy-Wilson. 2014.
\newblock Recognition of time-compressed speech does not predict recognition of
  natural fast-rate speech by older listeners.
\newblock \emph{The Journal of the Acoustical Society of America},
  136(4):EL268--EL274.

\bibitem[{Gu et~al.(2017)Gu, Neubig, Cho, and Li}]{Gu+:2017}
Jiatao Gu, Graham Neubig, Kyunghyun Cho, and Victor O.~K. Li. 2017.
\newblock \href {https://aclanthology.info/papers/E17-1099/e17-1099} {Learning
  to translate in real-time with neural machine translation}.
\newblock In \emph{Proceedings of the 15th Conference of the European Chapter
  of the Association for Computational Linguistics, {EACL} 2017, Valencia,
  Spain, April 3-7, 2017, Volume 1: Long Papers}, pages 1053--1062.

\bibitem[{Gumul(2006)}]{gumul2006conjunctive}
Ewa Gumul. 2006.
\newblock Conjunctive cohesion and the length of ear-voice span in simultaneous
  interpreting. a case of interpreting students.
\newblock \emph{Linguistica Silesiana}, (27):93--103.

\bibitem[{He et~al.(2016)He, He, Wu, and Wang}]{he+:2016}
Wei He, Zhongjun He, Hua Wu, and Haifeng Wang. 2016.
\newblock Improved neural machine translation with smt features.
\newblock In \emph{Proceedings of the Thirtieth AAAI Conference on Artificial
  Intelligence}, pages 151--157. AAAI Press.

\bibitem[{Imankulova et~al.(2019)Imankulova, Kaneko, Hirasawa, and
  Komachi}]{imankulova+:2019}
Aizhan Imankulova, Masahiro Kaneko, Tosho Hirasawa, and Mamoru Komachi. 2019.
\newblock Towards multimodal simultaneous neural machine translation.
\newblock \emph{arXiv preprint arXiv:2004.03180}.

\bibitem[{Inaguma et~al.(2020)Inaguma, Gaur, Lu, Li, and Gong}]{hirofumi+:2020}
Hirofumi Inaguma, Yashesh Gaur, Liang Lu, Jinyu Li, and Yifan Gong. 2020.
\newblock Minimum latency training strategies for streaming
  sequence-to-sequence asr.
\newblock \emph{ICASSP}.

\bibitem[{Lee(2002)}]{lee2002ear}
Tae-Hyung Lee. 2002.
\newblock Ear voice span in english into korean simultaneous interpretation.
\newblock \emph{M{\'e}ta: journal des traducteurs/M{\'e}ta: Translators'
  Journal}, 47(4):596--606.

\bibitem[{Lege(2012)}]{Lege2012TheEO}
Ryan Lege. 2012.
\newblock \emph{The Effect of Pause Duration on Intelligibility of Non-Native
  Spontaneous Oral Discourse}.
\newblock {PhD} dissertation, Brigham Young University.

\bibitem[{Li et~al.(2020)Li, Chang, Sainath, Pang, He, Strohman, and
  Wu}]{li+:2020}
Bo~Li, Shuo-Yiin Chang, Tara~N. Sainath, Ruoming Pang, Yanzhang He, Trevor
  Strohman, and Yonghui Wu. 2020.
\newblock Towards fast and accurate streaming end-to-end asr.

\bibitem[{Ma et~al.(2019)Ma, Huang, Xiong, Zheng, Liu, Zheng, Zhang, He, Liu,
  Li, Wu, and Wang}]{ma+:2019}
Mingbo Ma, Liang Huang, Hao Xiong, Renjie Zheng, Kaibo Liu, Baigong Zheng,
  Chuanqiang Zhang, Zhongjun He, Hairong Liu, Xing Li, Hua Wu, and Haifeng
  Wang. 2019.
\newblock \href {https://doi.org/10.18653/v1/P19-1289} {{STACL}: Simultaneous
  translation with implicit anticipation and controllable latency using
  prefix-to-prefix framework}.
\newblock In \emph{Proceedings of the 57th Annual Meeting of the Association
  for Computational Linguistics}, pages 3025--3036, Florence, Italy.
  Association for Computational Linguistics.

\bibitem[{Ma et~al.(2020{\natexlab{a}})Ma, Zheng, Liu, Zheng, Liu, Peng,
  Church, and Huang}]{ma2019incremental}
Mingbo Ma, Baigong Zheng, Kaibo Liu, Renjie Zheng, Hairong Liu, Kainan Peng,
  Kenneth Church, and Liang Huang. 2020{\natexlab{a}}.
\newblock Incremental text-to-speech synthesis with prefix-to-prefix framework.
\newblock \emph{Findings of EMNLP}.

\bibitem[{Ma et~al.(2020{\natexlab{b}})Ma, Pino, Cross, Puzon, and
  Gu}]{ma2019monotonic}
Xutai Ma, Juan Pino, James Cross, Liezl Puzon, and Jiatao Gu.
  2020{\natexlab{b}}.
\newblock Monotonic multihead attention.
\newblock \emph{8th International Conference on Learning Representations,
  {ICLR} 2020, Addis Ababa, Ethiopia, April 26-30, 2020, Conference Track
  Proceedings}.

\bibitem[{Oda et~al.(2014)Oda, Neubig, Sakti, Toda, and Nakamura}]{oda+:2014}
Yusuke Oda, Graham Neubig, Sakriani Sakti, Tomoki Toda, and Satoshi Nakamura.
  2014.
\newblock Optimizing segmentation strategies for simultaneous speech
  translation.
\newblock In \emph{ACL}.

\bibitem[{Oord et~al.(2017)Oord, Li, Babuschkin, Simonyan, Vinyals,
  Kavukcuoglu, Driessche, Lockhart, Cobo, Stimberg, Casagrande, Grewe, Noury,
  Dieleman, Elsen, Kalchbrenner, Zen, Graves, King, and Hassabis}]{oord+:2018}
Aaron Oord, Yazhe Li, Igor Babuschkin, Karen Simonyan, Oriol Vinyals, Koray
  Kavukcuoglu, George Driessche, Edward Lockhart, Luis Cobo, Florian Stimberg,
  Norman Casagrande, Dominik Grewe, Seb Noury, Sander Dieleman, Erich Elsen,
  Nal Kalchbrenner, Heiga Zen, Alex Graves, Helen King, and Demis Hassabis.
  2017.
\newblock Parallel {W}ave{N}et: Fast high-fidelity speech synthesis.
\newblock In \emph{ICML}.

\bibitem[{Ping et~al.(2017)Ping, Peng, Gibiansky, Arik, Kannan, Narang, Raiman,
  and Miller}]{ping+:2017}
Wei Ping, Kainan Peng, Andrew Gibiansky, Sercan~{\"O}mer Arik, Ajay Kannan,
  Sharan Narang, Jonathan Raiman, and John~L. Miller. 2017.
\newblock Deep {V}oice 3: Scaling text-to-speech with convolutional sequence
  learning.
\newblock In \emph{ICLR}.

\bibitem[{Sainath et~al.(2020)Sainath, He, Li, Narayanan, Pang, Bruguier,
  Chang, Li, Alvarez, Chen, Chiu, Garc{\'i}a, Gruenstein, Hu, Jin, Kannan,
  Liang, Mcgraw, Peyser, Prabhavalkar, Pundak, Rybach, Shangguan, Sheth,
  Strohman, Visontai, Wu, Zhang, and Zhao}]{sainath+:2020}
Tara~N. Sainath, Yanzhang He, Bo~Li, Arun Narayanan, Ruoming Pang, Antoine
  Bruguier, Shuo-Yiin Chang, Wei Li, Raziel Alvarez, Zhifeng Chen, Chung-Cheng
  Chiu, David Garc{\'i}a, Alex Gruenstein, Ke~Hu, Minho Jin, A.~Kannan, Qiao
  Liang, I.~Mcgraw, Cal Peyser, Rohit Prabhavalkar, Golan Pundak, David Rybach,
  Yuan Shangguan, Yash Sheth, Trevor Strohman, Mirk{\'o} Visontai, Yonghui Wu,
  Yinyong Zhang, and Ding Zhao. 2020.
\newblock A streaming on-device end-to-end model surpassing server-side
  conventional model quality and latency.
\newblock \emph{ICASSP}.

\bibitem[{Sennrich et~al.(2015)Sennrich, Haddow, and Birch}]{sennrich+:2015}
Rico Sennrich, Barry Haddow, and Alexandra Birch. 2015.
\newblock Neural machine translation of rare words with subword units.
\newblock \emph{arXiv preprint arXiv:1508.07909}.

\bibitem[{Shen et~al.(2018)Shen, Pang, Weiss, Schuster, Jaitly, Yang, Chen,
  Zhang, Wang, Skerrv-Ryan, Saurous, Agiomvrgiannakis, and Wu}]{shen+:2018}
Jonathan Shen, Ruoming Pang, Ron Weiss, Mike Schuster, Navdeep Jaitly, Zongheng
  Yang, Zhifeng Chen, Yu~Zhang, Yuxuan Wang, Rj~Skerrv-Ryan, Rif Saurous,
  Yannis Agiomvrgiannakis, and Yonghui Wu. 2018.
\newblock Natural {TTS} synthesis by conditioning {W}ave{N}et on {MEL}
  spectrogram predictions.
\newblock In \emph{Interspeech}.
\newblock Tacotron 2.

\bibitem[{Tachibana et~al.(2018)Tachibana, Uenoyama, and
  Aihara}]{tachibana+:2018}
Hideyuki Tachibana, Katsuya Uenoyama, and Shunsuke Aihara. 2018.
\newblock Efficiently trainable text-to-speech system based on deep
  convolutional networks with guided attention.
\newblock In \emph{2018 IEEE International Conference on Acoustics, Speech and
  Signal Processing (ICASSP)}, pages 4784--4788. IEEE.

\bibitem[{Vaswani et~al.(2017)Vaswani, Shazeer, Parmar, Uszkoreit, Jones,
  Gomez, Kaiser, and Polosukhin}]{vaswani+:2017}
Ashish Vaswani, Noam Shazeer, Niki Parmar, Jakob Uszkoreit, Llion Jones,
  Aidan~N Gomez, \L{}ukasz Kaiser, and Illia Polosukhin. 2017.
\newblock Attention is all you need.
\newblock In \emph{Advances in Neural Information Processing Systems 30}.

\bibitem[{Wang et~al.(2017)Wang, Skerry-Ryan, Stanton, Wu, Weiss, Jaitly, Yang,
  Xiao, Chen, Bengio, Le, Agiomyrgiannakis, Clark, and Saurous}]{wang+:2017}
Yuxuan Wang, RJ~Skerry-Ryan, Daisy Stanton, Yonghui Wu, Ron~J. Weiss, Navdeep
  Jaitly, Zongheng Yang, Ying Xiao, Zhifeng Chen, Samy Bengio, Quoc Le, Yannis
  Agiomyrgiannakis, Rob Clark, and Rif~A. Saurous. 2017.
\newblock \href {https://arxiv.org/abs/1703.10135} {Tacotron: Towards
  end-to-end speech synthesis}.
\newblock In \emph{Interspeech}.

\bibitem[{Xiong et~al.(2019)Xiong, Zhang, Zhang, He, Wu, and
  Wang}]{xiong+:2019}
Hao Xiong, Ruiqing Zhang, Chuanqiang Zhang, Zhongjun He, Hua Wu, and Haifeng
  Wang. 2019.
\newblock Dutongchuan: Context-aware translation model for simultaneous
  interpreting.

\bibitem[{Yamamoto et~al.(2020)Yamamoto, Song, and Kim}]{yamamoto+:19}
Ryuichi Yamamoto, Eunwoo Song, and Jae-Min Kim. 2020.
\newblock Parallel {W}ave{GAN}: {A} fast waveform generation model based on
  generative adversarial networks with multi-resolution spectrogram.
\newblock In \emph{ICASSP 2020-2020 IEEE International Conference on Acoustics,
  Speech and Signal Processing (ICASSP)}, pages 6199--6203. IEEE.

\bibitem[{Yuan and Liberman(2008)}]{yuan2008speaker}
Jiahong Yuan and Mark Liberman. 2008.
\newblock Speaker identification on the scotus corpus.
\newblock \emph{Journal of the Acoustical Society of America}, 123(5):3878.

\bibitem[{Zheng et~al.(2020{\natexlab{a}})Zheng, Liu, Zheng, Ma, Liu, and
  Huang}]{zheng+:2020}
Baigong Zheng, Kaibo Liu, Renjie Zheng, Mingbo Ma, Hairong Liu, and Liang
  Huang. 2020{\natexlab{a}}.
\newblock Simultaneous translation policies: from fixed to adaptive.
\newblock In \emph{Proceedings of the 58th Annual Meeting of the Association
  for Computational Linguistics}.

\bibitem[{Zheng et~al.(2019{\natexlab{a}})Zheng, Zheng, Ma, and
  Huang}]{zheng+:2019b}
Baigong Zheng, Renjie Zheng, Mingbo Ma, and Liang Huang. 2019{\natexlab{a}}.
\newblock Simpler and faster learning of adaptive policies for simultaneous
  translation.
\newblock In \emph{Proceedings of the 2019 Conference on Empirical Methods in
  Natural Language Processing and the 9th International Joint Conference on
  Natural Language Processing (EMNLP-IJCNLP)}, pages 1349--1354.

\bibitem[{Zheng et~al.(2019{\natexlab{b}})Zheng, Zheng, Ma, and
  Huang}]{zheng+:2019}
Baigong Zheng, Renjie Zheng, Mingbo Ma, and Liang Huang. 2019{\natexlab{b}}.
\newblock Simultaneous translation with flexible policy via restricted
  imitation learning.
\newblock In \emph{Proceedings of the 57th Annual Meeting of the Association
  for Computational Linguistics}, pages 5816--5822.

\bibitem[{Zheng et~al.(2019{\natexlab{c}})Zheng, Ma, Zheng, and
  Huang}]{zheng2019speculative}
Renjie Zheng, Mingbo Ma, Baigong Zheng, and Liang Huang. 2019{\natexlab{c}}.
\newblock Speculative beam search for simultaneous translation.
\newblock In \emph{Proceedings of the 2019 Conference on Empirical Methods in
  Natural Language Processing and the 9th International Joint Conference on
  Natural Language Processing (EMNLP-IJCNLP)}, pages 1395--1402.

\bibitem[{Zheng et~al.(2020{\natexlab{b}})Zheng, Ma, Zheng, Liu, and
  Huang}]{zheng+opportunistic:2020}
Renjie Zheng, Mingbo Ma, Baigong Zheng, Kaibo Liu, and Liang Huang.
  2020{\natexlab{b}}.
\newblock Opportunistic decoding with timely correction for simultaneous
  translation.
\newblock In \emph{ACL}.

\end{thebibliography}
\bibliographystyle{acl_natbib}

\end{CJK}
\end{document}